\documentclass[journal]{IEEEtran}
\ifCLASSINFOpdf
\else
\fi
\usepackage{amsmath, mathrsfs}

\usepackage{multirow, multicol, graphicx, amssymb, hyperref, booktabs, bm, hyperref}
\hypersetup{
    colorlinks=true,
    linkcolor=blue,
    filecolor=blue,      
    urlcolor=blue,
}
\usepackage[nomarkers]{endfloat}

\usepackage[switch,columnwise]{lineno}

\begin{document}
%
\title{A framework for deep learning emulation of numerical models with a case study in satellite remote sensing}
%
%
%

\author{Kate~Duffy,
        Thomas~Vandal,
        Weile~Wang,
        Ramakrishna~Nemani,
        and Auroop~R.~Ganguly
        
\thanks{K. Duffy and A. R. Ganguly are with the Sustainability and Data Sciences Laboratory, Department of Civil and Environmental Engineering, Northeastern University, 360 Huntington Avenue, Boston, MA 02115.}
\thanks{K. Duffy, T. Vandal, W. Wang, and R. Nemani are with NASA Ames Research Center, Moffett Blvd, Mountain View, CA 94035.}
\thanks{T. Vandal is with the Bay Area Environmental Research Institute, P.O. Box 25 Moffett Field, CA 94035.}
\thanks{W. Wang is with the California State University, Monterey Bay, Seaside, CA 93955.}
\thanks{Corresponding author: K. Duffy, duffy.k@northeastern.edu}
        
}

\maketitle

\begin{abstract}
Numerical models based on physics represent the state-of-the-art in earth system modeling and comprise our best tools for generating insights and predictions. Despite rapid growth in computational power, the perceived need for higher model resolutions overwhelms the latest-generation computers, reducing the ability of modelers to generate simulations for understanding parameter sensitivities and characterizing variability and uncertainty. Thus, surrogate models are often developed to capture the essential attributes of the full-blown numerical models. Recent successes of machine learning methods, especially deep learning, across many disciplines offer the possibility that complex nonlinear connectionist representations may be able to capture the underlying complex structures and nonlinear processes in earth systems. A difficult test for deep learning-based emulation, which refers to function approximation of numerical models, is to understand whether they can be comparable to traditional forms of surrogate models in terms of computational efficiency while simultaneously reproducing model results in a credible manner. A deep learning emulation that passes this test may be expected to perform even better than simple models with respect to capturing complex processes and spatiotemporal dependencies. Here we examine, with a case study in satellite-based remote sensing, the hypothesis that deep learning approaches can credibly represent the simulations from a surrogate model with comparable computational efficiency. Our results are encouraging in that the deep learning emulation reproduces the results with acceptable accuracy and often even faster performance. We discuss the broader implications of our results in light of the pace of improvements in high-performance implementations of deep learning as well as the growing desire for higher-resolution simulations in the earth sciences.
\end{abstract}

\begin{IEEEkeywords}
emulation, Bayesian deep learning, surrogate modeling, uncertainty quantification
\end{IEEEkeywords}

%
\IEEEpeerreviewmaketitle

\section{Introduction}
%
%
%
%

\IEEEPARstart{S}{atellite} remote sensing and climate modeling, which are our best tools for monitoring the climate and projecting the future climate, rely on powerful computers to process massive data streams and execute complex simulations. Over the last several decades, strides in high performance computing have enabled the development of climate models that solve equations at increasingly fine spatial resolutions. Past iterations of the Coupled Model Intercomparison Project (CMIP) have seen the average horizontal resolution decrease from 250km (CMIP3), to 150km (CMIP5), to 25-50km in the high-resolution experiments of the latest generation (CMIP6) \cite{ajibola2020evaluation}. Despite these strides, even the latest generation of climate models does not provide the type of robust regional climate information required by decisionmakers, and the extant literature suggests that high-resolution simulations of fine-scale processes are critical to reproducing extremes and change \cite{diffenbaugh2005fine}. Scientific and societal demand for higher-resolution simulations easily consume available resources, leaving less capacity to invest in model improvement runs such as parameters optimization and characterization of model sensitivity and variability. Lack of resources for science runs such as these may contribute to the relatively modest increases in historical skill and model consensus over time \cite{kumar2014regional, seferian2020tracking, mckenna2020indian}, despite major investments in climate modeling and computing.

In climate science, the only source of data growing faster than archived model simulations is remote sensing observations of the environment \cite{overpeck2011climate}.
According to the Global Climate Observing System, 26 out of 50 essential climate variables depend on satellite observations \cite{mason2007systematic}. Remotely sensed atmospheric and surface states document historical and current climates, are used to constrain the components of earth system models \cite{ghent2011data}, and are assimilated into climate reanalysis models \cite{saha2010ncep}.
Recent  decades  have  seen  an  ever-growing  fleet of Earth orbiting satellites and continuous resolution improvements. Depending on the spatial and temporal resolutions of gridded satellite datasets, as well as the number of attribute fields generated in post-processing, a low earth orbit satellite can generate tens to hundreds of gigabytes each year, while geostationary sensors like GOES-16 generate nearly one terabyte per day \cite{sun2019can, goes16}.

The cost of running complex models has motivated many studies of surrogate modeling in the earth sciences. Surrogate models capture the essential attributes of the full high-fidelity models, including their statistical variability, by projecting high-dimensional systems into low-order functions. These proxy models save orders of magnitude in computing time so that scientists can quickly study complex systems where repeated simulations would be prohibitively slow. For example, statistical modeling has been used to emulate the output of climate models under any arbitrary forcing scenarios, given a set of full models runs \cite{castruccio2014statistical, holden2010dimensionally}. Simplified approximations have also been evaluated as potential replacements for time-consuming portions of model physics \cite{krasnopolsky2005new}. In remote sensing, a range of emulation techniques have been proposed for radiative transfer models, which represent the scattering and absorption of solar radiation \cite{rivera2015emulator}. Even simpler reductions, such as lookup tables, are used operationally to reduce computation time \cite{maiac1}, while more complex emulators, such as those based on deep learning, have begun to show encouraging results \cite{kasim2020up}. 

Deep learning has transformed the state of the art in computer vision \cite{krizhevsky2017imagenet} natural language processing \cite{collobert2011natural}, and even complex games like chess and go \cite{gibney2016google}. 
In the earth sciences, deep learning has begun to show promise  for traditionally challenging problems such as El Niño/Southern Oscillation (ENSO) forecasting \cite{ham2019deep} and downscaling of geophysical data \cite{vandal2017deepsd, liu2020climate}.
It has been suggested that the ability to extract features from spatial and temporal context is at the root of deep learning’s promising results \cite{reichstein2019deep}, but
in the absence of of solid theoretical explanations for deep learning's success, performance improvements have been greeted as "unreasonable effectiveness" \cite{sejnowski2020unreasonable}. 
In scientific domains, where process understanding and and consistency with the laws of physics are key, "black box" machine learning models have not been universally embraced.
However, given that earth systems are governed by complex, multiscale, and nonlinear dynamics and contain spatial and temporal structures, several previous studies have begun to examine whether complex models may better capture the complexity of Earth science data than simpler, process-based models \cite{kratzert2019toward, lagerquist2021using}.

\begin{figure}[t]
\centering
\includegraphics[width=\textwidth]{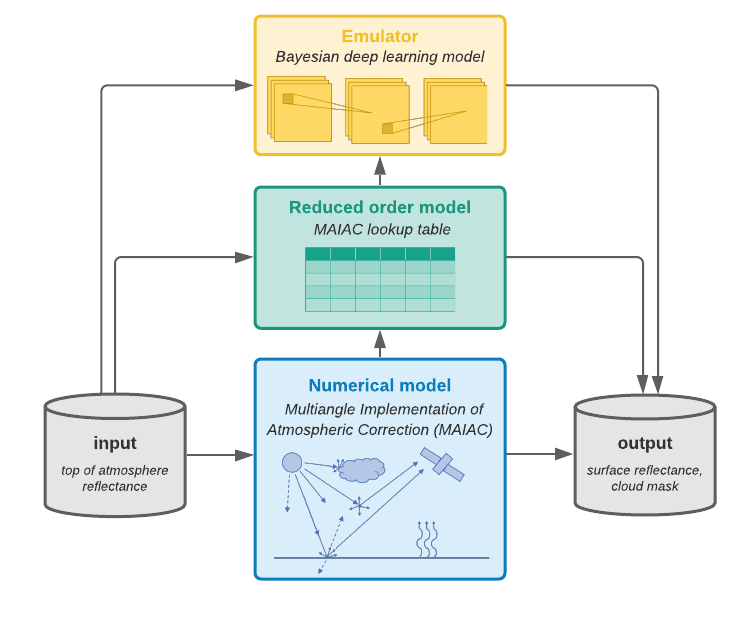}
\caption{Surrogate models save orders of magnitude in computing time so that scientists can quickly capture the essential attributes of complex simulations.
In the context of remote sensing, we examine the hypothesis that a deep learning emulator performs credibly and achieves similar computational performance when compared to a more traditional form of model order reduction.
}
\label{figure:overview}
\end{figure}

Here we present a case study comparing deep learning-based emulation to a traditional form of surrogate model in remote sensing.  A deep learning model is trained to approximate a lookup table capturing the radiative transfer model basis of the Multi-Angle Implementation of Atmospheric Correction (MAIAC) \cite{maiac1} algorithm using a dataset composed of lookup table inputs and outputs (Figure \ref{figure:overview}). We examine the hypothesis that deep learning approaches can credibly represent the simulations from a surrogate model and can achieve comparable or greater computational efficiency.

\section{Study Area and Data Sets}
Datasets used in this study are from the Advanced Himawari Imager (AHI) sensor carried by the Japanese geostationary satellite Himawari-8.
In the GeoNEX processing pipeline, Himawari Standard Data (HSD) scans are georeferenced and converted to gridded data. The resulting gridded data sets follow a geographic coordinate system with a 120$^{\circ}$ by 120$^{\circ}$ extent (E85$^{\circ}$ - E205$^{\circ}$, N60$^{\circ}$ - S60$^{\circ}$). The domain is divided into 6$^{\circ}$ by 6$^{\circ}$ tiles defined by fixed latitude and longitude.
Himawari-8’s full disk, which encompasses the entire view as seen from the satellite, covers the continent of Australia and eastern Asia. 
Full disk scans are repeated every 15 minutes throughout the day.

The Advanced Himawari Imager has sixteen observing bands encompassing visible, near-infrared (NIR), short wave infrared (SWIR) and thermal infrared with spatial resolution ranging from 0.5 to 2 km. Bands 1 through 6 are solar reflective bands, spectrally similar to NASA's Moderate Resolution Imaging Spectroradiometer (MODIS). 
All bands are resampled to common 0.01$^{\circ}$ resolution, which corresponds to 1 km at the equator [Table \ref{table:AHIbands}].

\begin{table}
\caption{Himwari-8 AHI solar reflective bands for land surface observation.}
\label{table:AHIbands}
\centering
\begin{tabular}{lcccccc}
\toprule
\textbf{Himwari-8 AHI Band}	& \textbf{Blue}	& \textbf{Green}	& \textbf{Red}	& \textbf{NIR}	& \textbf{SWIR1}	& \textbf{SWIR2}\\
\midrule
Center Wavelength ({\textmu}m) & 0.46 & 0.51 & 0.64 & 0.86 & 1.6 & 2.3 \\
Spatial resolution (km) & 1.0 & 1.0 & 1.0 & 0.5 & 2.0 & 2.0 \\
\bottomrule
\end{tabular}
\end{table}

\subsection{Himawari-8 AHI TOA Reflectance}
Top of atmosphere (TOA) reflectance is prepared by the GeoNEX processing pipeline from HSD scans, according to the processing procedure outlined in the Himawari 8/9 Himawari Standard Data User’s Guide, Version 1.2 \cite{himawari2015}. 
HSD consists of raw digital counts, which are transformed to Bidirectional Reflectance Factor (bands 1-6) and  Brightness Temperature (bands 7-16). 
TOA reflectance data and additional documentation are available from
\href{www.nasa.gov/geonex}{www.nasa.gov/geonex}.

\subsection{Himawari-8 AHI Surface Reflectance}
Originally developed for the land-monitoring flagship Moderate Resolution Imaging Spectrometer (MODIS), the MAIAC algorithm has been adapted to retrieve surface reflectance (SR) and atmospheric composition for the geostationary satellite Himawari-8.
Running an atmospheric radiative transfer model (RTM) for every measurement from a sensor is generally too slow and is not done in practice. 
Instead, MAIAC relies on the generation of look-up tables (LUT) from which values are retrieved by linear interpolation \cite{maiac1}.
As retrieval error is associated with step size, LUTs are precomputed at a grid density chosen with consideration to both accuracy and memory requirement.
Given the numerous possible variations in atmospheric conditions, keeping the LUT within reasonable size guides many choices made in its construction \cite{maiac1}.
In practice, the LUT approach gives nearly identical results as the inline RTM, given the necessary ancillary information such as sun-satellite geometries and atmospheric optical properties.

Geostationary surface reflectance is produced using the adapted MAIAC algorithm for all daylight observations \cite{li2019first}. 
MAIAC uses a time series of up to sixteen days and a mixture of pixel and image-level processing for atmospheric correction with internal cloud detection, aerosol retrieval and quality assurance flagging \cite{maiac1, maiac2, maiac3}. Multi-angle determination of surface reflectance refers to viewing angle, which is fixed, and illumination angle, which varies continuously throughout the day.
AHI MAIAC SR is released as a preliminary product and is available upon request at \href{www.nasa.gov/geonex}{www.nasa.gov/geonex}.

\subsection{MODIS MCD12Q1 Land Cover Type}
Land cover types are identified using MODIS global land cover classification, which is produced annually from combined Terra and Aqua observations \cite{sulla2018user}. 
MCD12Q1 incorporates five distinct classification schemes.
We use the International Geosphere Biosphere Programme (IGBP) global vegetation classification scheme.
This scheme delineates 17 distinct classes including 11 natural, 3 developed/mixed and 3 non-vegetated.
MCD12Q1 is resampled from 500 meter resolution to the 0.01 degree grid of the prepared AHI datasets.

\section{Methods}

\subsection{Emulator model}

Mapping from top of atmosphere to surface reflectance is relies on computationally expensive radiative transfer equations which simulate nonlinear physics and incorporate ancillary information about atmospheric conditions.
MAIAC is a state of the art method for accomplishing atmospheric correction and is run exclusively on an LUT, which achieves low ($\sim$0.2\%) error with respect to inline calculations.
Due to the unavailability of full model simulations in our domain of interest, in our approach to emulation, several deep networks are learned to approximate the input to output transformation contained in the MAIAC LUT.

\subsubsection{Bayesian Deep Learning}
\label{section:BDL}

Typical deep neural networks are learned as deterministic functions which fail to capture uncertainty in model parameters and data. The need to quantify these uncertainties has motivated development of approaches combining Bayesian probability theory and neural networks. 
Bayesian neural networks learn probability distributions over the network parameters, where training data is used to transform the prior probability distributions into posterior distributions.
Approximations of Bayesian inference have been postulated to extract information about both epistemic uncertainty (due to randomness) and aleatory uncertainty (due to incomplete systemic knowledge) from deep learning (DL) models \cite{kendall2017uncertainties}.

For a network with parameters $\mathbf{W}$, inputs $\mathbf{X}$ and outputs $\mathbf{Y}$, Bayesian theory estimates posterior distribution over the weights as $p(\mathbf{W}| \mathbf{X}, \mathbf{Y}) = p(\mathbf{Y}| \mathbf{X}, \mathbf{W}) p(\mathbf{W}) / p(\mathbf{Y}| \mathbf{X})$.
However, performing inference on the full posterior distribution is intractable for complex networks, where analytical solutions to the marginal probability $p(\mathbf{Y}| \mathbf{X})$ are unavailable \cite{graves2011practical}. 

This has led to the development of efficient approximations of Bayesian inference for deep learning \cite{wang2016towards}.
One approach to Bayesian deep learning defines an approximation of the true posterior distribution by Monte Carlo sampling.
Dropout, the process of randomly removing nodes from deep neural networks, is applied as a regularization technique in many deep learning models to discourage overfitting \cite{srivastava2014dropout}.
When applied during training and testing stochastic dropout generates thinned networks, sampling of which is mathematically equivalent to sampling the approximate posterior distribution. 

All models defined in this work are Bayesian models implemented with concrete dropout, a variant that dynamically adapts dropout probability to obtain well calibrated uncertainty estimates \cite{gal2017concrete}. 
For a model $f^\mathbf{W}(X)$  with $L$ layers, weights $\mathbf{W}$ and dropout probabilities $p$, the set of variational parameters is $\theta = \{\mathbf{w}_l, p_l\}_{l=1}^L$.
Thus, the  variational (factorized) approximation, $q_{\theta}$, of the true posterior distribution $p(\mathbf{W})$ is formalized as  $q_{\theta}(\mathbf{W}) = \prod_{l=1}^L q_{\theta}(\mathbf{w}_l)$.
The optimization objective $\mathscr{L}(\theta)$ for this variational interpretation of the model can be written as follows for data samples $\{x_i, y_i\}_{i=1}^N$ \cite{gal2017concrete}:


\begin{equation}
    \begin{split}
    \mathscr{L}(\theta) &= - \frac{1}{N} \sum_{i=1}^N log \: p(y_i |f^\mathbf{W}(\mathbf{x}_i))  + KL(q_\theta(\mathbf{W})||p(\mathbf{W})) \\
     &= \mathscr{L}_{\mathcal{X}}(\theta) + KL(q_\theta(\mathbf{W})||p(\mathbf{W}))
    \end{split}
\end{equation}
\\

\noindent The first term represents expresses the log likelihood of the model while Kullback–Leibler divergence ($KL$) term acts as a regularizer by discouraging separation between the approximate posterior and the true posterior distribution.
The loss function, $\mathscr{L}_{\mathcal{X}}(\theta)$  is written as follows for labels $\{y_i\}_{i=1}^N$ and output values $\{\hat{y}_i\}_{i=1}^N$:

\begin{equation}
\begin{aligned}
    \mathscr{L}_{\mathcal{X}}(\theta) = -\frac{1}{N} \sum_{i=1}^N \frac{1}{2} \hat{\sigma_i}^{-2} ||y_i - \hat{y_i}||^2 + \frac{1}{2}log \hat{\sigma_i}^2  \\
\end{aligned}
\end{equation}

\noindent Noise in the input data is represented using the uncertainty regularization term $\hat{\sigma_i}$ \cite{kendall2017uncertainties}. 
Through minimization of the loss function, the level of noise inherent in the data is learned implicitly.
Further, the model's epistemic uncertainty is characterized by the predictive variance \cite{kendall2017uncertainties}.
During inference, stochastic forward passes generate $T$ independent and identically distributed samples from $T$ thinned networks. From these samples, we can empirically approximate the model's predictive distribution at each pixel.
With $T$ samples of $[\hat{\mathbf{y}}, \hat{\sigma}]$ from the Bayesian network $f^\mathbf{W}(X)$ the unbiased estimates of the first two moments of the predictive distribution are:

\begin{equation}
E[y] = \frac{1}{T} \sum_{t=1}^T \hat{y}_t
\end{equation}

\begin{equation}
Var[y] = \frac{1}{T} \sum_{t=1}^T \hat{y}_t^2  - \biggl(\frac{1}{T} \sum_{t=1}^T \hat{y}_t \biggr)^2 + \sum_{t=1}^T \hat{\sigma}_t^2
\end{equation}

\noindent This estimate of uncertainty is obtained through supervision of the task rather than requiring ``uncertainty labels''.
It theoretically comprises both aleatoric uncertainty, from measurement noise, and epistemic uncertainty, reducible though collection of more data \cite{gal2016uncertainty}.

\subsubsection{Discrete-Continuous Distribution}
\label{section:DC}
Prediction tasks generally fall into one of two categories: 
regression tasks predict a continuous quantity, while
classification tasks are concerned with assigning a class label.
MAIAC's atmospheric correction and cloud classification algorithms generate surface reflectance, a continuous variable ranging between 0 and 1, as well as binary cloud classification. We learn a discrete-continuous model to perform both regression and classification tasks in one probabilistic model \cite{vandal2018quantifying}. To this end, the model is conditioned to predict the probability $\hat{p}$ of a pixel being clear sky. For the Bayesian network $f^\mathbf{W}(X)$ described in Section \ref{section:BDL}, the mean, variance and probability are sampled as follows:

\begin{equation}
[\hat{y}, \hat{\sigma}^2, \hat{\phi}] = f^\mathbf{W} (X) \\
\end{equation}
\begin{equation}
\hat{p} = Sigmoid(\hat{\phi})
\end{equation} 

This conditioning results in a two-part loss function with the first term capturing cross-entropy of predicted and cloud label and cloud prediction, and the second term capturing conditional regression loss at clear sky pixels. Here, the $y$ is the binary indicator of whether the classification is correct and $D$ is the number of pixels with pixel index $i$. The loss function in Equation 2 is modified as follows to incorporate classification loss:

\begin{equation}
\begin{aligned}
    \mathscr{L}_{\mathcal{X}}(\theta) =\overbrace{-\frac{1}{D} \sum_{i} \Big[ y_i\: log(\hat{p_i})) + (1-y_i)\:log(1-\hat{p_i})\Big] }^{binary \: classification \: loss} \\
    + \underbrace{\frac{1}{D} \sum_{i,y_i > 0}\frac{1}{2} \hat{\sigma_i}^{-2} ||y_i - \hat{y_i}||^2 + \frac{1}{2}log \hat{\sigma_i}^2}_{conditional \: regression \: loss}  \\
\end{aligned}
\end{equation}

\begin{figure}
\centering
\includegraphics[width=\textwidth]{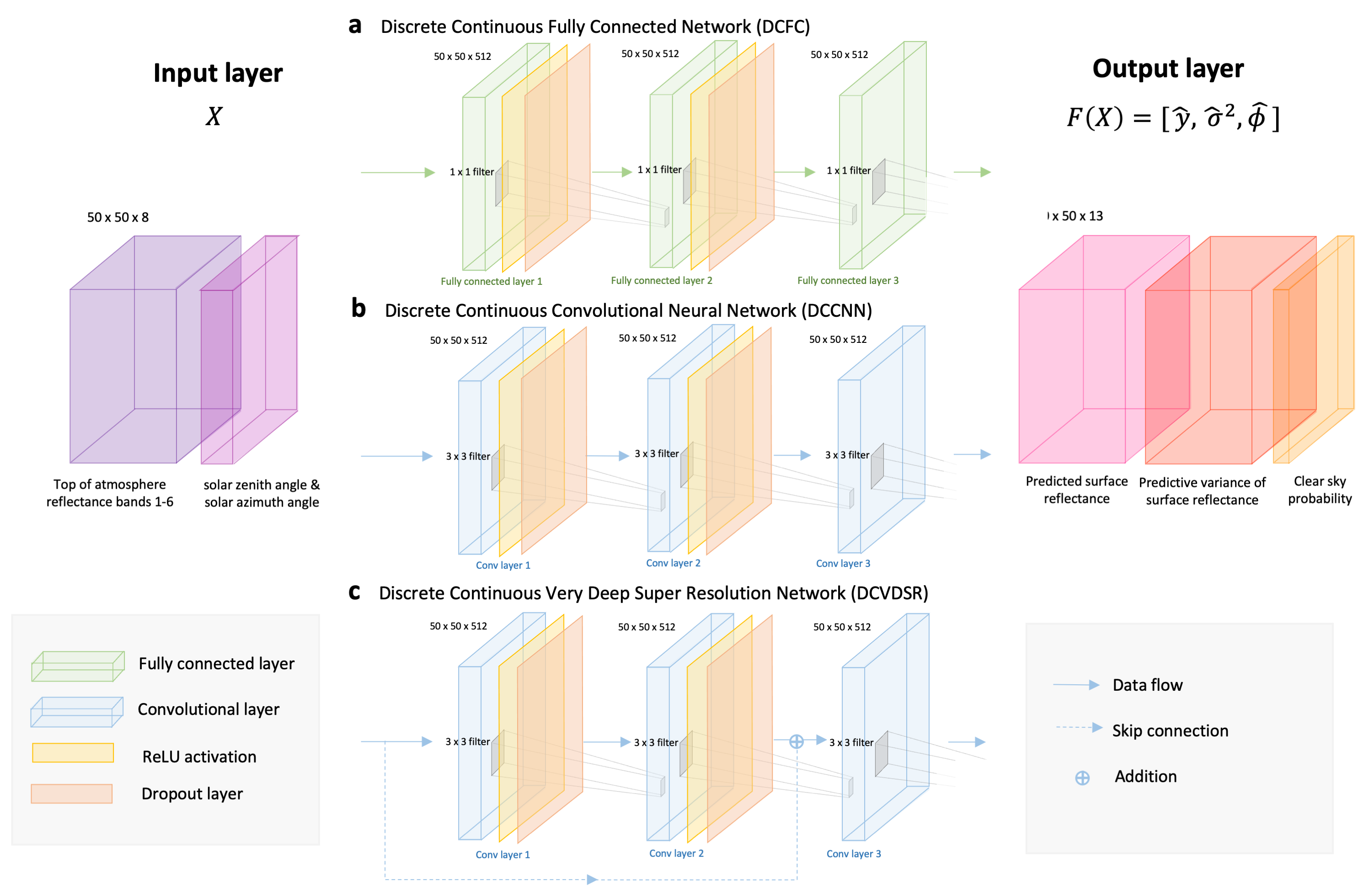}
\caption{Proposed model architectures for deep learning emulation. We implemented three full Bayesian architectures conditioned to learn a discrete-continuous distribution for surface reflectance and cloud mask prediction: \textbf{(a)} discrete-continuous fully connected neural network (DCFC), \textbf{(b)} discrete-continuous convolutional neural network (DCCNN), and  \textbf{(c)} Discrete-continuous very deep super resolution network (DCVDSR)} 
\label{figure:models}
\end{figure}

\subsubsection{Implementation and Training}
We implement three full Bayesian architectures conditioned to learn a discrete-continuous distribution as in Vandal et al. (2018) (Figure \ref{figure:models}).
The ML models are trained without the ancillary datasets (e.g. current atmosphere conditions) used as inputs to MAIAC.

The discrete-continuous fully connected neural network (DCFC) has three layers with 512 filters per layer.
The discrete-continuous convolutional neural network (DCCNN) is of similar width and depth to DCFC and has a convolutional filter size of three.
Finally, the discrete-continuous very deep super resolution network (DCVDSR), inspired by image super-resolution networks, is a convolutional neural network similar to DCCNN but incorporates a skip connection between the first and last hidden layers. The presence of the skip connection reformulates the task of the network as learning a residual function, or the difference between the output mapping and input \cite{he2016deep}. This reformulation improves performance by preconditioning the function as an identity mapping. 
Rather than adopting the common bottleneck type architecture, we opt to exclude pooling layers in all three designs. This choice is consistent with the task's dependence on relatively local information and helps to maintain temporal consistency in local features in the presence of global changes in the images.
All models use ReLU activations.

Over 60 TB of data from a two year period is divided into training (2016-01-01 to 2016-12-31) and testing (2017-01-01 to 2017-12-31) sets. The training set was used to experiment with model designs while the test set was used exclusively to evaluate the model and prepare figures.
Models are implemented in TensorFlow 2.0 and trained on 50 by 50 pixel patches for approximately eight hours using stochastic gradient descent and Adam optimization with $\beta_1=0.9$, $\beta_2=0.999$, $\epsilon=1e-7$, a batch size of 16, and learning rate of $1e-4$ ~\cite{kingma2014adam}.  
All three models use concrete dropout as described in Section \ref{section:BDL} for weight regularization and uncertainty quantification. For concrete dropout, hyperparameters tau and prior length-scale are set to $1e-5$ and $1e-14$.
Code is available at \href{https://github.com/KateDuffy/maiac-emulator}{github.com/KateDuffy/maiac-emulator}.

\subsection{Assessment of Emulated SR and Cloud Products}
Validation of surface reflectance products against in situ measurements would result in uncertainties due to the presence of mixed pixels at remote sensor's spatial resolution \cite{li2011retrieval, guo2020analysis}.
Therefore, comparison with an existing, comprehensively validated product can be used to assess the performance of a new reflectance product \cite{feng2012quality}.
Here, it is worthwhile to note that analysis and interpretation of the emulated data are performed in comparison to the lookup table prepared for MAIAC. 
These lookup tables are an approximation of a fundamentally approximate model and contain associated uncertainties \cite{maiac1}.
The task of the emulator can be posed as learning a smooth interpolation of the MAIAC lookup table.
As both the MAIAC SR and emulator SR are predicted from AHI TOA reflectance, all pixels are guaranteed coincident, coangled and colocated, and can be directly compared.
Additionally, we evaluate the ability of the emulator to discriminate between clear sky and non-clear sky pixels.
To evaluate stability of emulator performance under varied conditions, results are presented for performance for the nine common most MODIS land cover classifications and four seasons.

\begin{figure}[t]
\centering
\includegraphics[width=\textwidth]{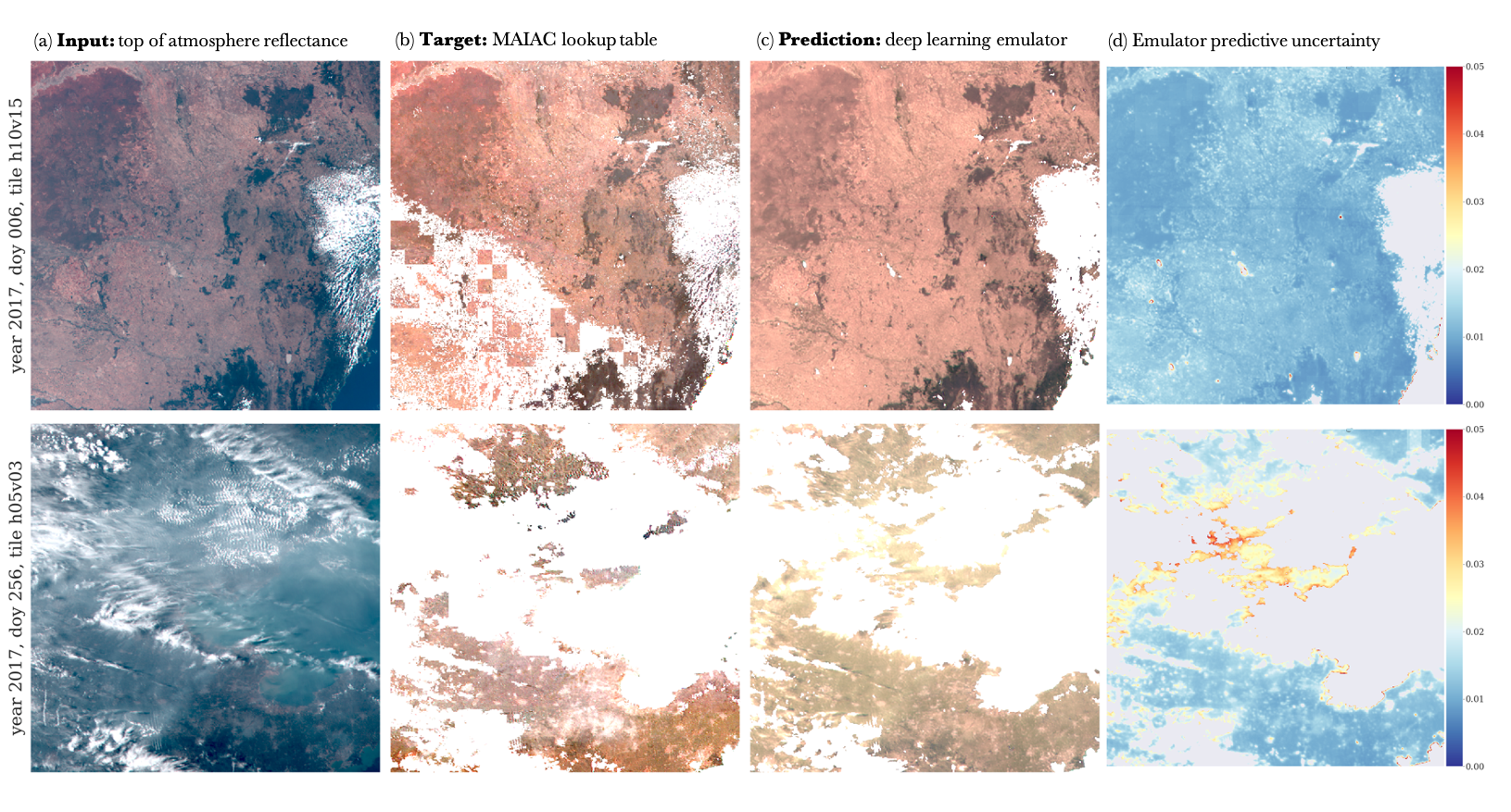}
\caption{
Correction of atmospheric effects and cloud masking performed on true color images from the Japanese geostationary satellite Himawari-8. \textbf{(a)} Top of atmosphere reflectance before processing. \textbf{(b)} Retrieval of surface reflectance and cloud mask (in white) produced by the surrogate MAIAC model. \textbf{(c)} Prediction of surface reflectance and cloud mask by the Bayesian deep learning emulator. \textbf{(d)} Visualization of predictive undertainty from the emulator model. \textbf{Rows from top:} southwestern Australia, eastern China and the Bohai Sea.
}
\label{figure:results}
\end{figure}

\section{Results and Discussion}

\subsection{Model Evaluation}

We adapt three deep learning architectures for comparison with MAIAC's surface reflectance and cloud retrieval lookup tables (Figure \ref{figure:results}). 
We compare the three models to identify the best-performing architecture, based on error in SR prediction, accuracy in cloud identification and uncertainty quantification.
Most in-depth analysis is focused on DCVDSR, which is the best-performing model in uncertainty calibration.

\subsubsection{Surface Reflectance}
A comparison between basic statistics of surface reflectance datasets obtained from MAIAC and the emulators is presented in Table \ref{table:SRresults}. 
Here, the mean value of surface reflectance is the average intensity of clear sky pixels in each band. 
Coefficient of variation (CV) is a measure of relative dispersion of the data calculated as the ratio between the standard deviation and mean of a distribution. 
For surface reflectance, CV relates to the radiometric stability characteristic of the sensor, with lower CV indicating greater stability.
Comparison of MAIAC and emulator CV suggest that the predictions of the fully connected model (DCFC) most closely match the dispersion of MAIAC SR.
The emulator models generally capture the relative magnitudes of variation in each wavelength while underestimating variation of the SR distribution. Underestimation of the observed variation is most pronounced for the blue band across models.

\begin{figure}
\centering
\includegraphics[width=\textwidth]{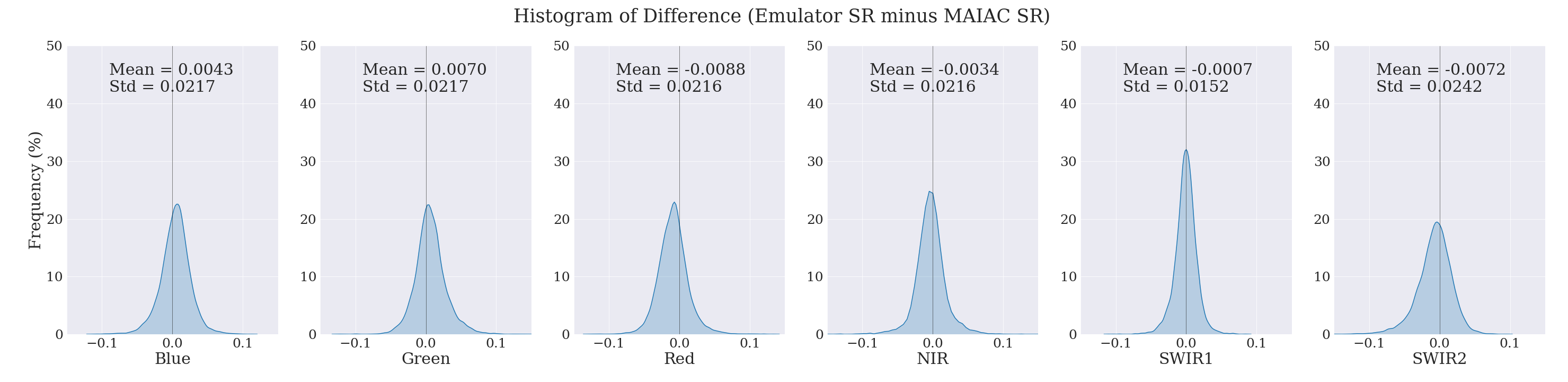}
\caption{Histograms of difference between MAIAC lookup table surface reflectance and emulator (DCVDSR) surface reflectance for six bands indicate minimal bias toward over- or underestimation of surface reflectance by the emulator.}
\label{figure:histograms}
\end{figure}

Histograms of differences between MAIAC and emulator surface reflectance are plotted in Figure \ref{figure:histograms}.
For an ideal model, differences between observed and modeled values should be small and unbiased. 
Distributions are generally symmetric and centered around near-zero means, indicating minimal bias toward overestimation or underestimation by the emulator. Similar results are obtained for the three candidate emulator models.


Correlation coefficient and conditional root mean square error (RMSE) are also presented in Table \ref{table:SRresults}. Conditional RMSE refers to RMSE evaluated at the pixels identified by both MAIAC and the emulator as clear sky. 
Correlation coefficients indicate the strongest linear relationship between MAIAC SR and DCFC emulator SR.
This result is unsurprising as the convolutional filters in DCCNN and DCVDSR will produce smoother results than the pixelwise model, DCFC.
These correlation coefficients reflect similar strength of linear relationship as those between MAIAC SR and MODIS SR \cite{li2019first}. MODIS Terra and Aqua products provided the principle comparison for validation of AHI MAIAC SR.
Evaluation of the performance across models suggests that mappings in some bands may be easier to learn (SWIR1, SWIR2), and others more difficult to learn (Blue, Green).
Conditional RMSE is generally lowest for the DCVDSR model.
As the square root of the variance of residuals, RMSE indicates the absolute fit of the model to the data, and can be thought of as more germane to predictive ability than correlation.
In the context of the accuracy requirements of higher level data products (Albedo: 0.02-0.05, NDVI: 0.03) \cite{vermote2006surface} the emulators generally err within reasonable range.

\begin{table}
\caption{Evaluation of surface reflectance from the three candidate emulator models. 
}
\label{table:SRresults}
\centering
\begin{tabular}{l|l|cccccc}
\toprule
& & \textbf{Blue}	& \textbf{Green}	& \textbf{Red}	& \textbf{NIR}	& \textbf{SWIR1}	& \textbf{SWIR2}\\
\midrule
\multirow{1}{*}{$\overline{SR}$} 
&  MAIAC & 0.064  & 0.084 & 0.155 & 0.307 & 0.327 & 0.231 \\
\midrule
\multirow{3}{*}{$\overline{SR}_{MAIAC} - \overline{SR}_{emulator}$} 
& DCFC & \textbf{-0.001} & \textbf{-0.003} & -0.015 & -0.012  & -0.012 & -0.031  \\
& DCCNN & 0.002 & 0.006 & \textbf{0.008} & 0.012  & 0.014 & \textbf{0.006} \\
& DCVDSR & -0.004 & -0.009 & 0.010 & \textbf{0.005} & \textbf{0.002} & 0.008 \\
\midrule
\multirow{1}{*}{CV (\%)} 
& MAIAC & 88 & 80 & 62 & 34  & 38 & 51 \\
\midrule
\multirow{3}{*}{$\frac{CV_{MAIAC} - CV_{emulator}}{CV_{MAIAC}}$ (\%)}
& DCFC & \textbf{9.1} & \textbf{4.2} & \textbf{2.3} & 6.8  & 3.7 & \textbf{1.0} \\
& DCCNN & 31 & 24  & 10 & \textbf{1.7} & \textbf{-0.84} & 4.1  \\
& DCVDSR & 30 & 11 & 12 & 11  & 3.4 & 4.8 \\
\midrule
\multirow{3}{*}{Correlation coefficient (R)} 
& DCFC & \textbf{0.960} & \textbf{0.962} & 0.842 & 0.893 & \textbf{0.984} & \textbf{0.963} \\
& DCCNN & 0.948 & 0.950 & 0.818 & 0.866 & 0.977 & 0.960   \\
& DCVDSR & 0.792 & 0.860 & \textbf{0.930} & \textbf{0.918} & 0.977 & 0.956 \\
\midrule
\multirow{3}{*}{Conditional RMSE} 
& DCFC & 0.0221 & 0.0217 & 0.0235 & \textbf{0.0196}  & 0.0186 & 0.0387 \\
& DCCNN & 0.0228 & 0.0239 & 0.0228 & 0.0228  & 0.0228 & 0.0235 \\
& DCVDSR & \textbf{0.0211} & \textbf{0.0210} & \textbf{0.0220} & 0.0211  & \textbf{0.0153} & 0\textbf{.0233} \\
\bottomrule
\end{tabular}
\end{table}

\begin{figure}
\centering
\includegraphics[width=\textwidth]{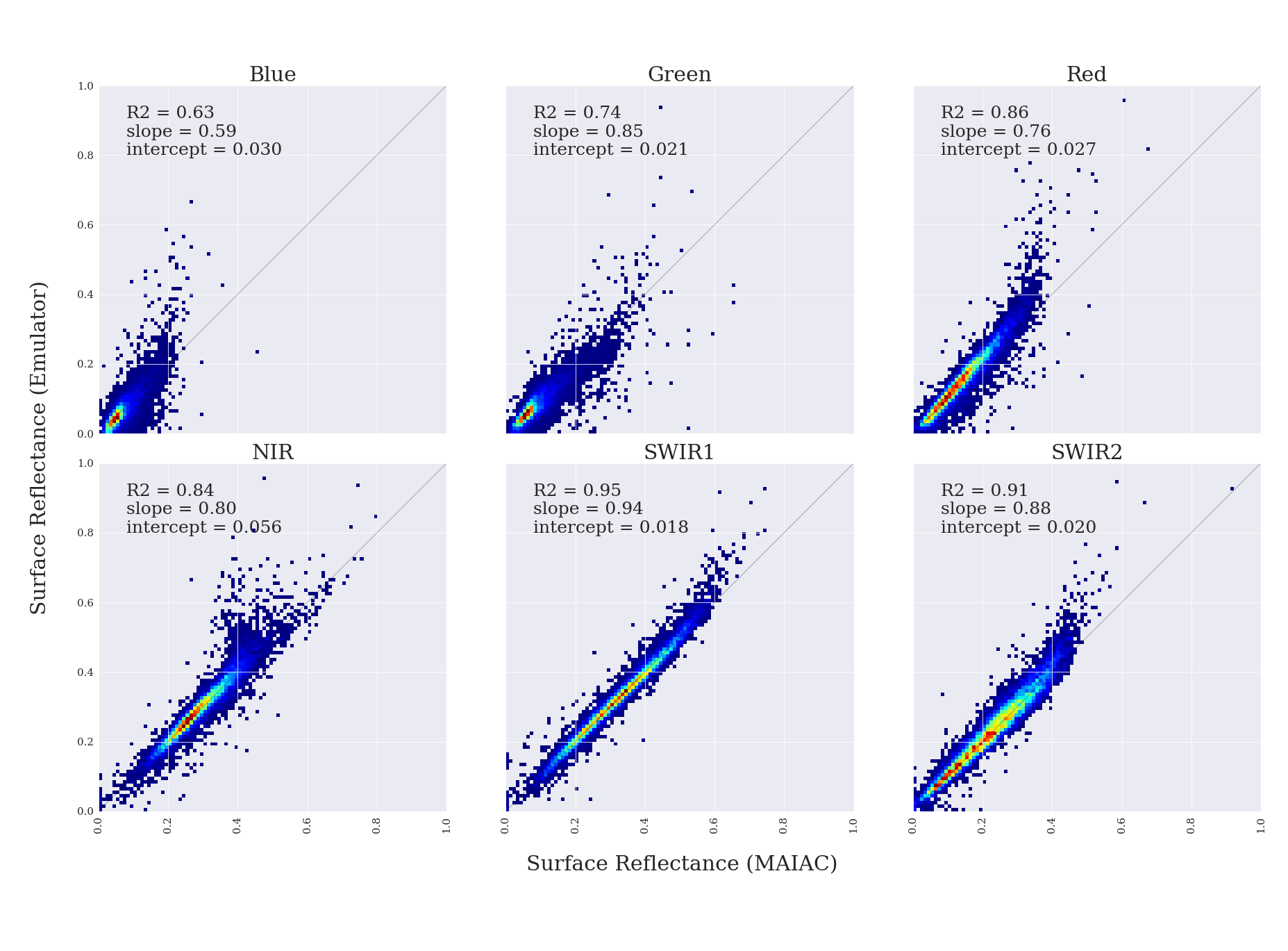} 
\caption{Density scatter plots compare surface reflectance from the MAIAC lookup table and the deep learning emulator (DCVDSR) for the six Advanced Himawari Imager solar reflective bands. Red indicates a higher density of points and blue a lower density. The gray line is a 1:1 line.}
\label{figure:R2plot}
\end{figure}

Pixel by pixel comparison is presented using density plots in Figure \ref{figure:R2plot}.
The plots suggest strong coherence between MAIAC and emulator SR.
A 1:1 line is displayed for visual comparison, while slope and intercept of the data best fit line are displayed on the plots.
Outliers are generally located above the 1:1 line, suggesting that the emulator model does not fully capture the upper tail of the SR distribution.

\subsubsection{Cloud Identification}
Evaluation of the emulator cloud prediction is performed by pixelwise comparison between the MAIAC cloud mask and the emulator cloud mask.
Binary classification accuracy is defined as the fraction of true predictions, where $TP$ = true positives, $TN$ = true negatives, $FP$ = false positives and $FN$ = false negatives.

\begin{equation}
    \begin{aligned}
    Accuracy = \frac{TP + TN}{TP + TN + FP + FN}
    \end{aligned}
\end{equation}
\\

\begin{figure}
\centering
\includegraphics[width=\textwidth]{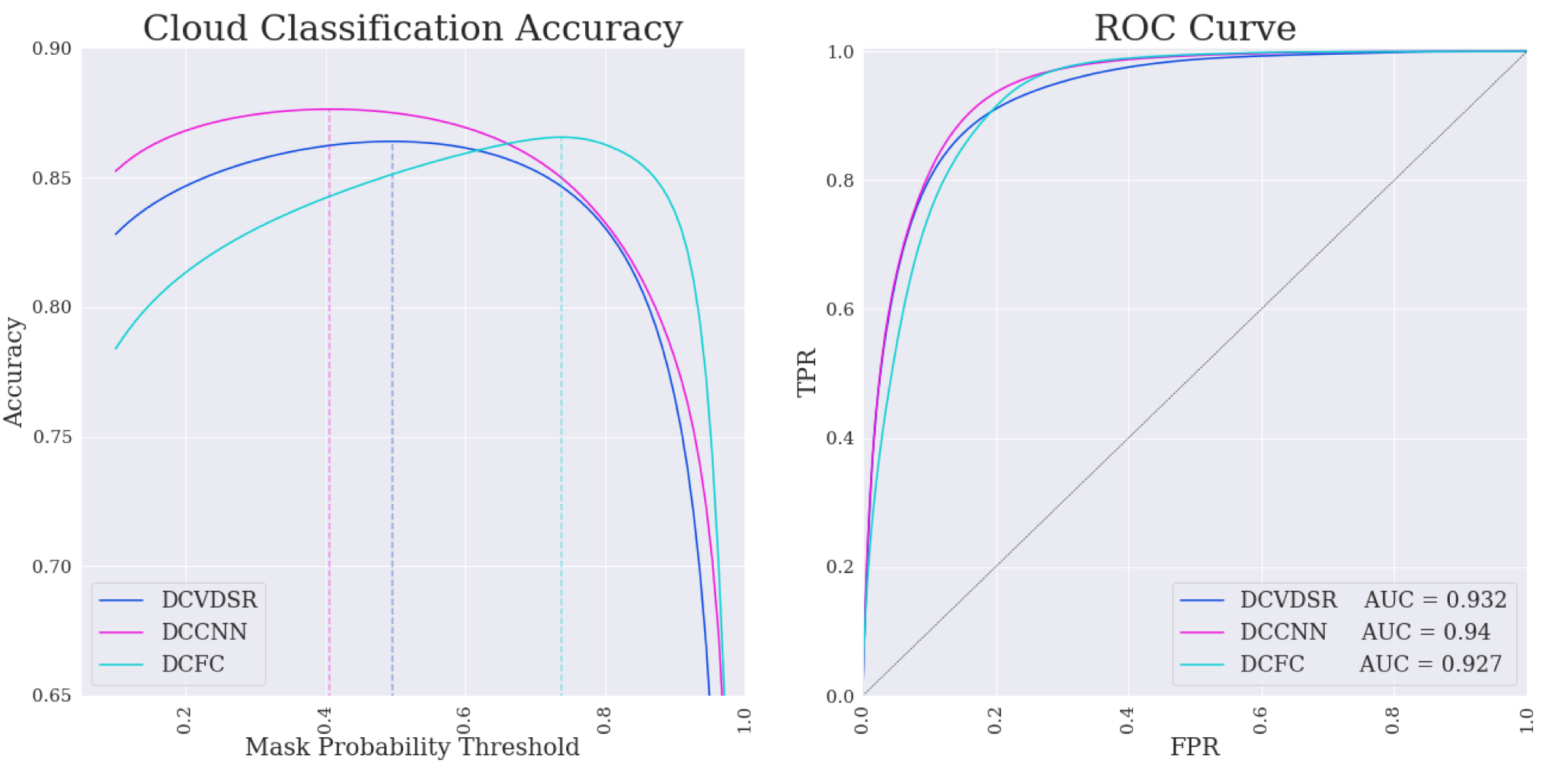}
\caption{Analysis of cloud prediction by the deep learning emulators. \textbf{(a)} Cloud classification accuracy with varying decisions thresholds under the discrete-continuous distribution. Models achieve 86 to 87\% accuracy when compared to MAIAC lookup table cloud masks. Dotted lines represent the optimal threshold for each model. \textbf{(b)} Area under the ROC curve for evaluation of the model's discrimination ability, with an area under the curve (AUC) value of 1 denoting perfect discrimination, and the gray 1:1 line representing no skill.} 
\label{figure:cloud}
\end{figure}

As described in Section \ref{section:DC}, the model is conditioned to predict $\hat{p}$ as the probability of a pixel being clear sky.
By selecting a decision threshold value of $\textbf{p}$, cloud classification proceeds by casting pixels with $\hat{p}<\textbf{p}$ as non-clear sky and $\hat{p}>\textbf{p}$ as clear sky.
Continuously varying the decision threshold $\textbf{p}$ and calculating the resulting classification accuracy indicates the optimal mask probability, $\textbf{p}$, for each trained model.
Figure \ref{figure:cloud} presents a plot of classification accuracy with varying decision threshold.

Figure \ref{figure:cloud} also presents the receiver operating characteristic (ROC) curve, used to assess discrimination ability of binary classifiers. 
True positive rate (TPR) is plotted against false positive rate (FPR) at various thresholds. 
Area under the curve provides a threshold-invariant measure of how well the model can discriminate between two classes, with a maximum value of one for perfect classification.

Performance by accuracy for cloud classification is similar across the evaluated emulator models (Table \ref{table:cloud}).
Sensitivity refers to the true positive rate, or proportion of clear sky pixels that are correctly classified.
Specificity refers to the true negative rate,  or the proportion of non-clear pixels that are correctly classified.
A high specificity classifier will erroneously screen high aerosol optical depth pixels, while a less conservative, higher sensitivity classifier carries more chance of cloud contamination. 
Such cloud contamination has a potentially strong negative effect on SR retrieval.
The three emulator models are generally conservative, achieving greater classification accuracy for non-clear pixels than clear sky pixels.

\begin{table}
\caption{Cloud classification accuracy, sensitivity and specificity.}
\label{table:cloud}
\centering
\begin{tabular}{l|cccc}
\toprule
& \textbf{Accuracy (\%)}	& \textbf{Sensitivity}	& \textbf{Specificity}\\
\midrule
DCFC & 0.866 & 0.702 & \textbf{0.921} \\
DCCNN & \textbf{0.877} & 0.767 & 0.905 \\
DCVDSR & 0.864 & \textbf{0.771} & 0.872 \\
\bottomrule
\end{tabular}
\end{table}

It should be noted that assessment of classification accuracy uses MAIAC cloud masks as ground truth. 
Cloud masks produced from MAIAC contain uncertainties and inaccuracies of their own, and it is possible that the ability of CNNs to incorporate spatial information produces an advantage in cloud classification.
Visual assessment of cloud predictions from MAIAC and emulator often indicate greater spatial coherence of emulator cloud masks, and lesser appearance of some undesirable model artifacts (Figure \ref{figure:results}).

\subsubsection{Stability of Model over Varied Conditions}

\begin{figure}
\centering
\includegraphics[width=\textwidth]{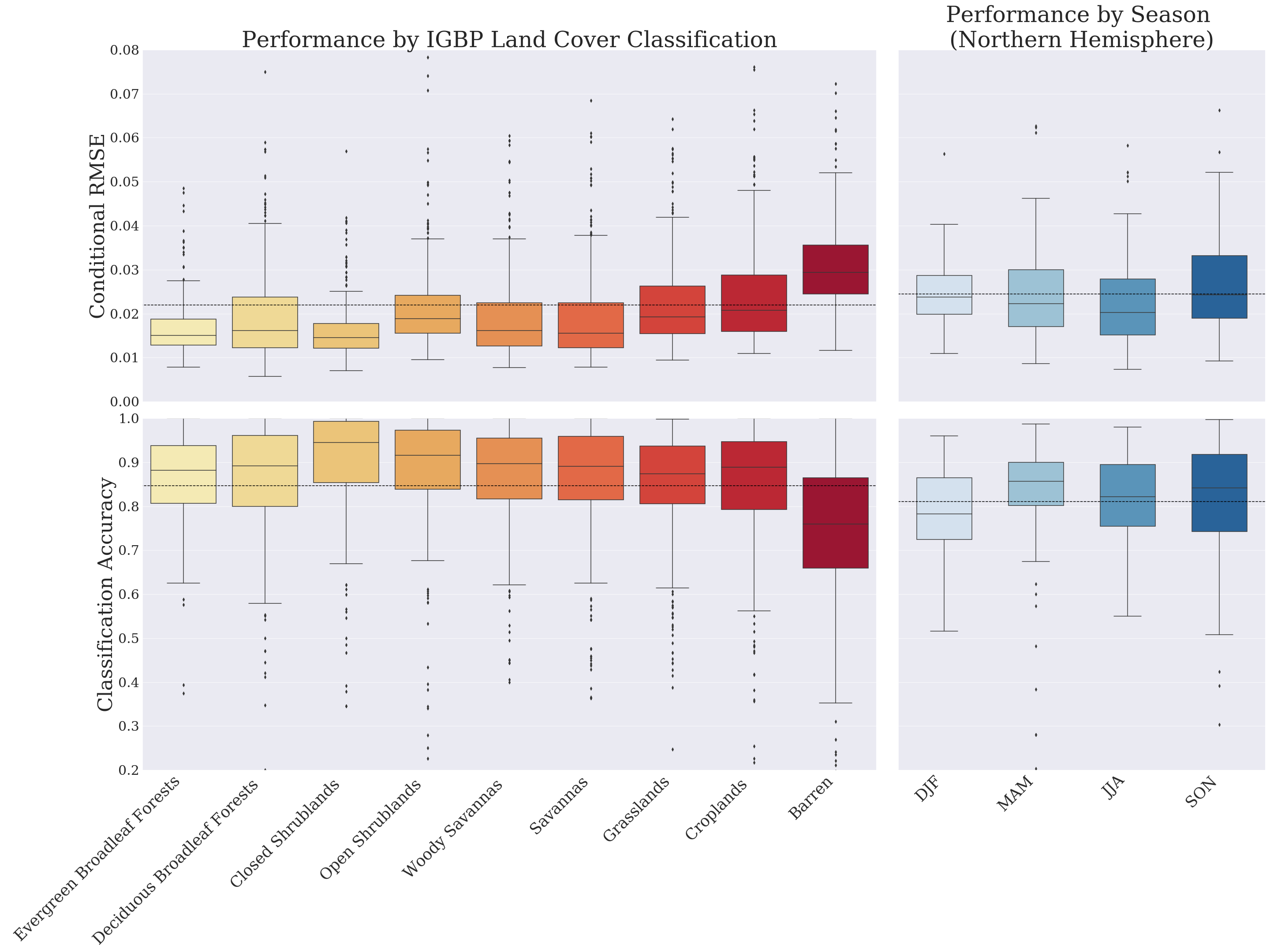}
\caption{Analysis of emulator surface reflectance error and cloud classification accuracy across varying surface conditions. \textbf{(a)} Emulator (DCVDSR) performance is relatively stable across vegetated land types, according to the IGBP global vegetation classification scheme. 
\textbf{(b)} Emulator (DCVDSR) performance in the Northern Hemisphere is slightly poorer in winter than in seasons. 
Dotted lines represent overall mean performance for all land covers and all seasons in the Northern Hemisphere, respectively.}
\label{figure:landcover}
\end{figure}

Homogeneous vegetation areas are identified using MODIS MCD12Q1 Land Cover Type I and performance of the MAIAC emulator is evaluated for each land cover type separately.
Performance including conditional RMSE of SR and cloud classification accuracy by the DCVDSR emulator are presented in Figure \ref{figure:landcover}.
Results are presented for the nine most abundant classes in the test set.
Both regression error and cloud classification accuracy are relatively stable for vegetated categories including forests, shrubland and savanna, while barren land results in poorer performance.
The optical properties of highly reflective surfaces present a challenge to atmospheric correction, and such may also result in poor performance for MAIAC \cite{lyapustin2012improved}. 
In addition to high reflectance, large solar angle, large zenith angle, and high aersol optical depth also form challenging conditions for reflectance retrieval.
More specific evaluations are needed to understand the performance of the emulator under these special conditions.

Seasonal analysis is used to evaluate the performance under annual fluctuations in vegetation phenology (Figure \ref{figure:landcover}).
Spring green-up, fall senescence and transitions between wet and dry seasons result in SR variation of several absolute percent in vegetated areas \cite{maiac3}.
SR error and cloud classification accuracy are generally stable throughout the year, but evince slightly poorer performance and greater spread fall months (SR prediction) and winter months (cloud classification).
Seasonal performance is evaluated separately for each hemisphere for consistency of seasons.
Similar results were found for the Southern Hemisphere.

\subsection{Uncertainty Quantification}

Bayesian deep learning models capture predictive uncertainty in regression task by producing a probabilistic output.
As described in Section \ref{section:BDL},  we use variational inference to produce an ensemble of predictions for each sample, then compute unbiased estimates of the first and second moments of the predictive distribution at each pixel.
From the second moment, the standard deviation expresses the magnitude of predictive uncertainty at each location.

\begin{figure}
\centering
\includegraphics[width=0.48\textwidth]{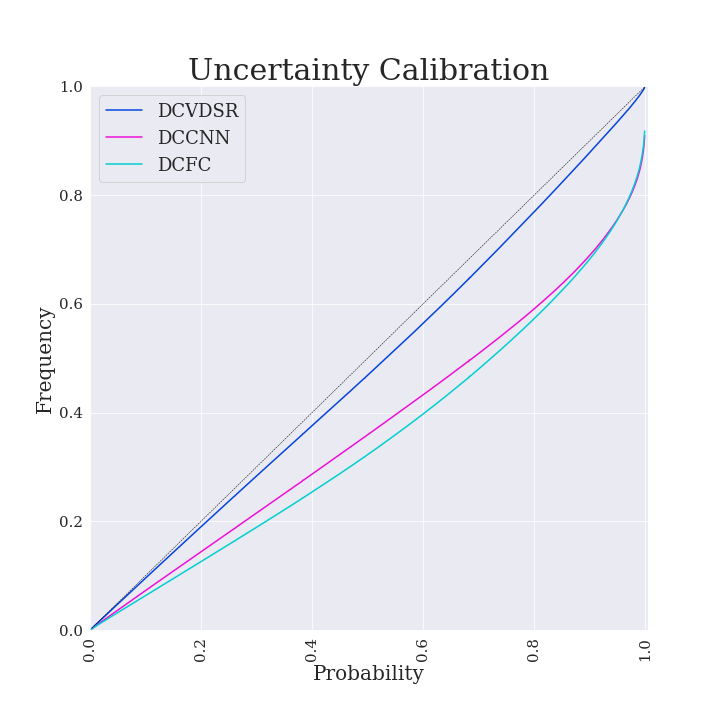}
\caption{
Assessment of predictive uncertainty across the three Bayesian deep learning emulator models.
Uncertainty calibration, which evaluates the frequency of observed values (y-axis) within predicted probability ranges (x-axis), indicates underestimation of uncertainty to varying extents. The gray 1:1 line indicates the behavior of a perfectly calibrated model.}
\label{figure:UQ}
\end{figure}

We assess the quality of the uncertainty measurements by evaluating the uncertainty calibration, or whether the model captures the uncertainty in observed data.
We compare the model's predictive distribution to the observed values by evaluating the frequency of residuals lying in various probability thresholds within the predicted distribution \cite{kendall2017uncertainties}
(Figure \ref{figure:UQ}). A perfectly calibrated model, which captures the distribution of the observed data, would match the 1:1 line.
All three models underestimate uncertainty to some extent, meaning they are overconfident in their predictions.
Of the three, DCVDSR has most well-calibrated uncertainty.

While quantification of uncertainty in deep learning makes a step toward more informed use of artificial intelligence models,  interpretability remains a key challenge in earth science and other domains. Alternatives to dropout, such as Bayes by Backprop, have exhibited comparable performance  in recent years \cite{blundell2015weight}. Further theoretical development and experimentation with practical applications will improve the trustworthiness of AI.

\subsection{Computational Performance}
The spatiotemporal resolution and spatial extent of AHI scans result in the generation of 94 GB of new data per day. 
There is growing interest in using AI to exploit this data, as well as the growing volume from other remote sensors, climate reanalysis models, and earth system models.

We quantify the nontrivial computing time necessary to retrieve surface reflectance using the MAIAC lookup table and the emulator models.
To evaluate the deep learning emulator models, we assess inference from one forward pass (static network) and ten stochastic forward passes (Bayesian sampling network).
A single inference with the static network is sufficient to produce SR and cloud products; Bayesian sampling produces the same with uncertainty quantification. Processing speeds are presented in Table \ref{table:speed}.
Emulator inference is evaluated on one GPU, while
the MAIAC lookup tables are executed on one CPU.
Among the compared emulator models, processing speed decreases with increasing complexity.
Inference with Bayesian sampling is generally comparable in speed to the LUT, while inference on the static network represents between 3.75x (DCDVSR) and 6x (DCFC) speedup over the acceleration offered by the LUT.

GPU-based models also have advantages in scalability. 
Parallel workloads distributed across GPUs can achieve a speedup which is nearly linear with the number of cores, while scaling across additional CPU cores lags in performance.
These behaviors will tend to increase the advantages of GPU-based computing in higher throughput operations. Additionally, advances in distributed computing and performance-aware implementations of deep learning, while outside the scope of this work, will further enhance efficiency. 
Meanwhile improvements in the implementation of MAIAC and the design of lookup tables provide a moving target for both statistical and computational performance.

\begin{table}
\caption{Processing speed of MAIAC and emulator models.}
\label{table:speed}
\centering
\begin{tabular}{l|c|c}
\toprule
\multirow{3}{*}{Model} & \multicolumn{2}{c}{Examples per second} \\
\cline{2-3}
& \multirow{2}{*}{Static network} & Bayesian sampling \\
 & & network \\
\midrule
MAIAC LUT & 0.40 & --- \\
DCFC & 2.4 & 0.60  \\
DCCNN & 1.8 & 0.33  \\
DCVDSR & 1.5 & 0.25  \\
\bottomrule
\end{tabular}
\end{table}

\section{Conclusions}
The time and expense associated with running numerical models often create a bottleneck in research on complex systems.
In this work we present a case study evaluating deep learning-based emulation in comparison to a more traditional reduced order approximation of the MAIAC model.
We found initially promising results in the statistical and computational performance of the deep learning emulator.
However, in counterbalance to deep learning's unfettered success in many commercial domains, we encountered challenges unique to working with geoscience data.
Satellite observations exhibit significant heterogeneity in space and time, with the joint distributions between variables fluctuating across locations and seasons. We mitigated this effect by using smaller patches to allow for larger batch sizes with more heterogeneity of samples.
And whereas any form of model reduction represents a trade-off between model fidelity and speed, the "black box" nature of deep learning models is a particular disadvantage in the geosciences, where physical consistency and interpretability are key.
Appropriate adaptations of recent work in explainable and interpretable deep learning \cite{toms2019physically}, as well as physics-guided neural networks \cite{karpatne2017physics}, may offer greater reliability and new research directions in the context of emulation.
For example, networks that trace salience back to the input information can be used to infer scientifically meaningful information about what the model has learned.
Future directions should also include further evaluation of the model's ability to generalize, especially under rare or challenging conditions.

As computation time is the main motivating factor in this work, it is important to consider the multiple factors contributing to the outlook on emulations's computational advantages.
Just as high-performance implementations of deep learning and GPU hardware are activate areas of research, MAIAC and its reduced order implementations are not static but are under ongoing development.
Advances in numerical modeling and CPU-based computing will alter the landscape across many scientific disciplines and shift the benchmark for speedups.
As such, we anticipate a framework for deep learning emulation to evolve continuously with new generations of hardware and software technology.
Our current results suggest that as advances continue in machine learning and high performance computing, the credibility and computational performance of deep learning emulators may improve sufficiently to ease computing-limitations on model development and big data processing across multiple scientific disciplines.




\ifCLASSOPTIONcaptionsoff
  \newpage
\fi



%
\clearpage
\bibliographystyle{ieeetr}
\bibliography{bib}

%








\end{document}